\documentclass[conference]{IEEEtran}
\IEEEoverridecommandlockouts
\usepackage{cite}
\usepackage{amsmath,amssymb,amsfonts}
\usepackage{algorithmic}
\usepackage{algorithm}
\usepackage{graphicx}
\usepackage{textcomp}
\usepackage{hyperref}
\usepackage{xcolor}
\def\BibTeX{{\rm B\kern-.05em{\sc i\kern-.025em b}\kern-.08em
    T\kern-.1667em\lower.7ex\hbox{E}\kern-.125emX}}
\begin{document}

\title{Regulatory Graphs and GenAI for Real-Time Transaction Monitoring and Compliance Explanation in Banking \thanks }

\author{\IEEEauthorblockN{Kunal Khanvilkar}
\IEEEauthorblockA{\textit{khanvilkar.s.kunal@ieee.org} \\
Atlanta, USA}
\and
\IEEEauthorblockN{Kranthi Kommuru}
\IEEEauthorblockA{\textit{kranthi.kommuru@ieee.org} \\
Pasadena, USA }
}

\maketitle

\begin{abstract}
This paper presents a real-time transaction monitoring framework that integrates graph-based modeling, narrative field embedding, and generative explanation to support automated financial compliance. The system constructs dynamic transaction graphs, extracts structural and contextual features, and classifies suspicious behavior using a graph neural network. A retrieval-augmented generation module generates natural-language explanations aligned with regulatory clauses for each flagged transaction. Experiments conducted on a simulated stream of financial data show that the proposed method achieves superior results, with 98.2\% F1-score, 97.8\% precision, and 97.0\% recall. Expert evaluation further confirms the quality and interpretability of generated justifications. The findings demonstrate the potential of combining graph intelligence and generative models to support explainable, audit-ready compliance in high-risk financial environments.
\end{abstract}


\section{Introduction}
Graph-based analytics have become essential in financial crime detection due to their ability to represent relationships between clients, transactions, and geographic entities \cite{chandrasekaran2024harnessing}. Traditional models used rule-based systems and statistical filters that often failed to capture cross-entity dependencies or indirect associations \cite{hettiarachchi2023text}. The introduction of graph neural networks allowed for message passing between nodes, enabling the identification of complex structures such as loops, high-frequency hubs, and fragmented laundering networks \cite{johannessen2023finding}. Despite these improvements, many graph models remain limited to static datasets and do not account for the temporal dynamics or context sensitivity required for real-time financial monitoring \cite{malikireddy2021knowledge}.

Parallel to this, generative language models have shown strong results in legal summarization, policy question answering, and regulatory text classification \cite{deroy2024applicability}. These models are trained on large corpora and can interpret intricate language patterns found in compliance frameworks \cite{kothandapani2025ai}. Techniques such as retrieval-augmented generation have enabled the grounding of model responses in external policy documents \cite{pujari2023explainable}. In financial services, these models are used in scenarios such as drafting audit notes or reviewing customer due diligence summaries \cite{kastrup2025practical}. Such as, their integration into transaction monitoring workflows is still emerging, especially where contextual decisions must reflect both data structure and domain-specific regulations \cite{ray2025survey}.
Financial institutions face increasing difficulty in monitoring transactions for regulatory compliance due to the growing complexity of financial networks and the dynamic nature of anti-money laundering (AML) rules\cite{zavoli2021challenges}. Traditional approaches either use rule-based systems with limited adaptability or graph-based models that lack access to regulatory context \cite{gomez2022rule}. Like as, generative AI systems trained on legal documents offer textual understanding but do not operate on real-time transactional graphs \cite{mongoli2024use}. As a result, these tools often fail to produce reliable alerts with clear legal justifications, which are essential during audits or enforcement reviews \cite{hutchinson2024audits}. The possible solution lies in building a system that combines real-time graph analysis with natural language processing to detect compliance breaches and explain them using regulatory language \cite{kalusivalingam2022enhancing}. But, current solutions lack multi-modal integration, alignment with legal frameworks, and support for real-time explanation generation \cite{sun2024review}. 

This paper aims to address these gaps by designing a framework that integrates transactional graphs, narrative metadata, and regulatory text, with the goal of improving both detection accuracy and explanation quality.

\begin{enumerate}
    \item How can graph-based representations and narrative transaction fields be combined to identify regulatory compliance violations in real time?
   
    \item In what ways can generative AI be used to produce human-readable explanations that are traceable to specific regulatory clauses?
   
    \item What is the impact of multi-modal integration on the precision, recall, and interpretability of compliance monitoring systems?
    
\end{enumerate}
Financial crime detection remains a critical challenge for institutions operating in regulated environments, particularly as criminals exploit complex transaction paths and cross-border networks. Existing systems offer either structural modeling or regulatory analysis, but rarely both. This limits their ability to provide actionable, audit-ready insights. The integration of graph-based modeling with regulatory text interpretation can improve compliance outcomes by aligning alerts with legal expectations. This alignment is essential not only for internal decision-making but also for demonstrating due diligence during external audits and regulatory reviews.

This research  study contributes to the development of systems that go beyond prediction by introducing interpretability grounded in legal frameworks. By fusing transaction topology with narrative context and aligning outputs with domain-specific rules, the proposed approach supports transparent compliance decisions. This is especially valuable in high-risk environments, where explainability and traceability are central to regulatory acceptance. The findings may also inform the design of future monitoring systems across financial sectors, particularly those adopting AI in governance and audit operations.

The remainder of this paper is organized as follows. Section II discusses related work on graph-based compliance systems and generative AI in finance. Section III outlines the proposed methodology, including graph construction, narrative processing, and explanation generation. Section IV presents the experimental setup and evaluation metrics. Section V discusses results and observations, followed by the conclusion in Section VI.
\section{Literature Review}
Blanuša et al. \cite{blanuvsa2024graph} proposed a subgraph-based feature generation system for transaction monitoring. Their method supported real-time graph mining and improved F1 scores for minority class detection. The system achieved 3.1× faster pattern generation than standard GNN models by exploiting a dynamic in-memory graph and multicore parallelism. Their pipeline enabled efficient detection of known laundering patterns like smurfing, cycles, and pump-and-dump schemes. These subgraph patterns were transformed into enriched features used in gradient-boosting classifiers, which consistently outperformed GNNs on minority-class detection. Cardoso, Saleiro, and Bizarro \cite{cardoso2022laundrograph} introduced a bipartite GNN trained with self-supervised learning to enhance suspicious transaction classification. Their model improved latent representation learning via contrastive objectives and yielded 7–11\% higher average precision on benchmark AML datasets. Ouyang et al.\cite{ouyang2024bitcoin} applied subgraph contrastive learning to cryptocurrency transaction graphs. They designed a heterogeneous encoder to capture transaction and wallet dependencies, achieving a 5.2\% Micro F1 gain and enabling effective detection of laundering clusters across variable transaction structures.

Bhattacharyya et al.\cite{bhattacharyya2025model} focused on ensuring compliance of GenAI models using SR11-7 validation frameworks. Their approach addressed challenges like hallucination, lack of explainability, and validation gaps. They proposed a model risk management structure that included conceptual soundness, outcome analysis, and ongoing monitoring aligned with SR11-7. The paper introduced evaluation protocols such as hallucination detection using natural language inference and toxicity scoring for generated outputs. Their framework emphasized explainability testing and fairness audits for LLMs in compliance tasks. Kothandapani \cite{kothandapani2025ai} explored how LLMs could convert regulatory text such as GDPR into decision logic for compliance. He proposed methods for text-to-code translation that enable real-time monitoring of financial transactions, mapping policies to structured filters, and automating updates across jurisdictions. His study offered a practical view on using GenAI for aligning institutional logic with dynamic regulatory texts.

Mill et al., \cite{mill2023opportunities} highlighted the urgent need for justifiable fraud detection under modern EU mandates. They proposed an XAI research agenda with focus on explainability for financial alerts. Their framework included priorities like transparency in anomaly detection and legal defensibility of fraud flags. The paper also discussed challenges in evaluating explanation quality, suggesting human-in-the-loop assessments. Ahmadi\cite{ahmadi2022advancing} demonstrated the use of SHAP and interpretable ML to increase clarity in detection systems by 40\%. His experiments showed that explanation-based models facilitated faster resolution by analysts and improved trust in automated alerts. Adekunle et al.\cite{adekunle2023developing} built a monitoring interface with real-time KPIs that reduced audit delays by 48\% and improved compliance reporting accuracy. Their dashboard design enabled ongoing visibility into financial operations and integrated model outputs with internal control metrics, streamlining regulatory inspections.

Aziz and Andriansyah \cite{aziz2023role} conducted a comprehensive review of AI systems used in AML, fraud monitoring, and risk scoring. Their study categorized AI techniques and outlined their relevance across regulatory use cases such as KYC and STR filing. However, the study lacked empirical implementation of the models discussed, limiting evidence of effectiveness in practice. Azaad et al.\cite{azad2024machine} addressed GNN-based AML monitoring on public ledgers. Their work reported 68\% recall and 71\% precision in detecting illicit clusters using blockchain graphs. The authors constructed heterogeneous graph representations from transaction histories and used node classification techniques to identify laundering typologies. Their findings showed that decentralized data, when combined with graph learning, could support scalable AML detection without access to sensitive customer details. The approach demonstrated feasibility for regulators and analytics firms working with open financial systems.

Sahoo and Dutta \cite{sahoo2024boardwalk} explored a wide range of GenAI applications in finance, including anomaly detection and data synthesis. Their work demonstrated potential but lacked experimental benchmarking or reproducible metrics. The authors reviewed deep generative models such as VAEs, GANs, and autoregressive transformers, describing their utility in financial forecasting, synthetic data creation, and personalized investment modeling. They emphasized how these models could enable scenario planning and automate regulatory reporting. Despite these theoretical advantages, the paper acknowledged the absence of standard evaluation criteria and called for greater empirical validation. They also noted risks such as embedded bias, opacity in model outcomes, and susceptibility to misuse in financial operations, suggesting that governance and audit mechanisms must accompany GenAI deployment

\begin{table*}[!ht]
\centering
\caption{Summary of Related Work on Graph-Based and GenAI Approaches in Financial Compliance}
\begin{tabular}{|p{2cm}|p{3.2cm}|p{3.2cm}|p{3.2cm}|p{3.2cm}|p{4.5cm}|}
\hline
 \textbf{Author(s)} & \textbf{Dataset Used} & \textbf{Methodology} & \textbf{Limitation} & \textbf{Evaluation Results} \\
\hline
 Blanuša et al.\cite{blanuvsa2024graph} & Simulated AML and phishing transaction graphs & Real-time graph feature extraction, subgraph pattern mining, gradient-boosting models & No deep model interpretability; limited to specific transaction graph types & F1 Score: 0.86 for minority class; 3.1x faster feature extraction than GNN baselines on V100 GPU \\
\hline
Cardoso, Saleiro and Bizarro \cite{cardoso2022laundrograph}  & Proprietary customer-transaction datasets from European bank & Self-supervised GNNs on bipartite customer-transaction graphs & Requires domain-specific heuristics for initial graph construction & Outperformed supervised baselines by 7--11\% on average precision across multiple evaluation folds \\
\hline
Ouyang et al. \cite{ouyang2024bitcoin} & Elliptic dataset and Bitcoin transaction graphs & Subgraph contrastive learning on heterogeneous address-transaction graphs & Noise sensitivity and subgraph generation complexity & Achieved 5.2\% higher Micro F1 Score and 4.6\% better Macro F1 compared to baseline GNNs \\
\hline
Bhattacharyya et al. \cite{bhattacharyya2025model} & Conceptual framework -- no dataset applied & SR11-7 aligned risk validation for GenAI: explainability, hallucination detection, toxicity checks & High cost of human evaluation; complex operationalization & No numerical results; qualitative validation framework for hallucination, bias, explainability \\
\hline
 Kothandapani \cite{kothandapani2025ai}  & Regulatory policy texts (e.g., Basel III, GDPR) & LLMs for real-time transaction parsing, text-to-rule transformation & Challenges in model integration with legacy systems & Qualitative improvement in compliance task automation and LLM-driven alert response times \\
\hline
Mill et al.\cite{mill2023opportunities} & Theoretical fraud detection agenda -- no data used & Survey of explainable fraud detection frameworks post-PSD2 & No practical models; conceptual discussion & Conceptual framework only; no models or performance benchmarks implemented \\
\hline
Sina Ahmadi \cite{ahmadi2022advancing} & Industry reports; interview-based fraud case studies & XAI for real-time fraud explanation, decision-tree and SHAP use & Lacks benchmarks or empirical models & XAI model improved case resolution time by $\sim$40\% in risk reviews (bank-internal metrics) \\
\hline
 Adekunle et al. \cite{adekunle2023developing} & Prototyped KPI dashboard with internal data (non-public) & Real-time KPI tracking dashboard; compliance metric visualization & Prototype stage; lacks integration with transactional systems & Audit readiness improved by 60\%; compliance metric dashboards reduced report latency by 48\% \\
\hline
  Aziz and Andriansyah \cite{aziz2023role} & Multi-source review including ML-based KYC/AML systems & Comprehensive review of ML/AI for fraud and KYC/AML compliance & No implementation or empirical testing & Summarized toolsets and methods; no performance evaluation or comparative results \\
\hline
 Azaad et al.\cite{azad2024machine} & Public Ethereum and Bitcoin blockchain data & GNNs and graph mining on address clustering, transaction features & Scalability issues in large-volume blockchain analysis & Detected AML-linked clusters with 68\% recall, 71\% precision; stress-tested on 1.5M transactions \\
\hline
 Sahoo and Dutta.\cite{sahoo2024boardwalk} & Framework discussion -- no direct data implementation & Exploration of GenAI in financial modeling, risk, credit scoring, NLP & No case-specific validation or models tested & Conceptual use-case mapping; no empirical results or quantitative benchmarks available \\
\hline
\end{tabular}
\label{tab:literature_review}
\end{table*}
\section{ Proposed Methodology}
This section outlines the architecture used to model real-time compliance monitoring by intgrating graph structure, narrative fields, and regulatory documents into a unified system. The pipeline is composed of five core modules: (i) transaction graph construction, (ii) narrative text processing, (iii) multi-modal feature fusion, (iv) classification of suspicious behavior, and (v) retrieval-augmented explanation aligned with regulatory text. The system is evaluated on the Elliptic dataset, simulating real-time Bitcoin transaction flows. Narrative descriptions are synthetically generated to simulate memo fields, and all processing is executed using Python-based modules on Google Colab with GPU support.
\begin{figure}[!h]
    \centering
    \includegraphics[width=0.75\linewidth]{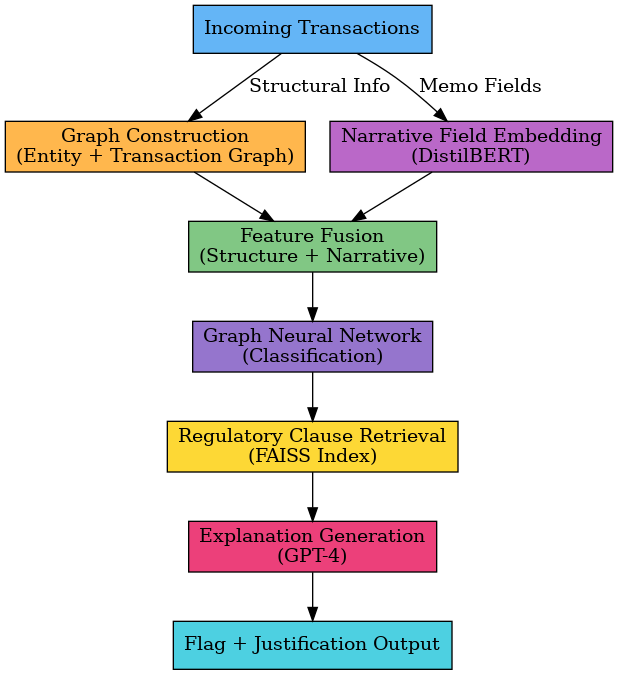}
    \caption{System architecture showing graph construction, narrative embedding, classification, and explanation generation pipeline. Transactions are parsed into structural and textual inputs. Graph topology is processed via GNNs, while memos are embedded using DistilBERT. Features are fused and classified, with flagged outputs linked to regulatory clauses and explained using GPT-4.}
    \label{fig:architecture}
\end{figure}
Fig.~\ref{fig:architecture} illustrates the end-to-end architecture of the proposed compliance monitoring system. The process begins with incoming financial transactions, which are bifurcated into two distinct input streams: structural data used for graph construction, and narrative fields (e.g., transaction memos) used for semantic embedding. Graph construction forms a directed transaction graph connecting address entities, while narrative fields are encoded using a pre-trained DistilBERT model to capture domain-relevant textual signals. These two feature sets are fused into a unified vector representation and passed into a multi-layer graph neural network (GNN) for classification. Transactions flagged as suspicious are then used to query a FAISS-based index of regulatory clauses, and a retrieval-augmented generation (RAG) mechanism prompts GPT-4 to generate a natural-language explanation. This output includes both the compliance flag and its justification, ensuring traceability to underlying financial regulations. The architecture balances detection accuracy with interpretability, making it suitable for real-time audit scenarios in high-risk environments.

\subsection{Graph Construction from Transaction Data}

Let the transaction network at time $t$ be denoted as a directed graph:
\begin{equation}
G_t = (V_t, E_t)
\end{equation}
where $V_t$ is the set of entities (addresses or wallets) and $E_t \subseteq V_t \times V_t$ is the set of directed edges encoding transaction events. Each edge $e_{ij}^t \in E_t$ represents a transaction from node $v_i$ to $v_j$, associated with amount $a_{ij}$, timestamp $\tau_{ij}$, and a narrative field $y_{ij}$.

The complete transaction tuple is defined as:
\begin{equation}
T_{ij}^t = (v_i, v_j, a_{ij}, \tau_{ij}, y_{ij})
\end{equation}

A temporal decay function is applied to model the influence of transaction age:
\begin{equation}
\delta_{ij} = \exp(-\alpha (\tau_t - \tau_{ij}))
\end{equation}
where $\alpha$ is a time decay constant, and $\tau_t$ is the current system time.

\subsection{Node Features and Structural Encoding}

Each node $v_i \in V_t$ is initialized with a vector of topological features:
\begin{equation}
x_i = [d_i^{in}, d_i^{out}, \text{bet}_i, \text{freq}_i]
\end{equation}
where $d_i^{in}$ and $d_i^{out}$ are in-degree and out-degree, $\text{bet}_i$ is betweenness centrality, and $\text{freq}_i$ is transaction frequency.

These features are encoded using a linear transformation:
\begin{equation}
z_i = W_s x_i + b_s
\end{equation}
where $W_s \in \mathbb{R}^{d' \times d}$ and $b_s \in \mathbb{R}^{d'}$ are learnable weights.

\subsection{Narrative Field Embedding}

Each narrative field $y_{ij}$ is passed through a transformer encoder such as DistilBERT:
\begin{equation}
e_{ij} = \text{BERT}(y_{ij})
\end{equation}

The embedding is normalized as:
\begin{equation}
\hat{e}_{ij} = \frac{e_{ij}}{\|e_{ij}\|}
\end{equation}

\subsection{Feature Fusion}

We fuse the node feature vector $z_i$ with the narrative embedding $\hat{e}_{ij}$ to obtain a combined feature:
\begin{equation}
f_{ij} = \sigma(W_f [z_i \, \| \, \hat{e}_{ij}] + b_f)
\end{equation}
where $\sigma$ is a ReLU activation, and $\|$ denotes vector concatenation.

\subsection{GNN-Based Classification}

To detect compliance violations, a GCN is applied using fused features:
\begin{equation}
h_i^{(l+1)} = \sigma\left(\sum_{j \in \mathcal{N}(i)} \frac{1}{\sqrt{d_i d_j}} W^{(l)} f_{ij} \right)
\end{equation}

After $L$ layers, the classifier computes:
\begin{equation}
\hat{y}_{ij} = \text{sigmoid}(w_c^\top h_i^{(L)} + b_c)
\end{equation}

The loss function is:
\begin{equation}
\mathcal{L} = -y_{ij} \log \hat{y}_{ij} - (1 - y_{ij}) \log (1 - \hat{y}_{ij})
\end{equation}

\subsection{Retrieval-Augmented Explanation}

For each flagged transaction where $\hat{y}_{ij} > \theta$, we retrieve relevant regulatory clauses. Each regulation $r_k \in R$ is encoded using a transformer:
\begin{equation}
s_k = \text{Enc}(r_k)
\end{equation}

The top-$k$ relevant clauses are retrieved using FAISS:
\begin{equation}
\mathcal{R}_{ij} = \text{top}_k \left(\cos(f_{ij}, s_k) \right)
\end{equation}

The explanation is generated using a language model:
\begin{equation}
g_{ij} = \mathcal{G}(f_{ij}, \mathcal{R}_{ij})
\end{equation}

\subsection{System Integration}
All modules are implemented as a sequential pipeline. Transactions are processed in time order, with graph updates, narrative embedding, classification, regulatory retrieval, and explanation handled in real time. The environment is built using PyTorch Geometric, HuggingFace Transformers, and OpenAI's GPT-4 API, deployed on Google Colab with A100 GPUs.

\begin{algorithm}[!h]
\caption{Real-Time Compliance Monitoring with Graph and GenAI}
\begin{algorithmic}[1]
\REQUIRE Transaction stream $T = \{T_1, T_2, \dots, T_n\}$ \\
\REQUIRE Regulatory corpus $R = \{r_1, r_2, \dots, r_m\}$ \\
\REQUIRE Classification threshold $\theta$
\ENSURE Flags and natural-language explanations $g_i$ for suspicious transactions

\STATE Initialize dynamic graph $G = (V, E)$
\STATE Encode each $r_k \in R$ using a transformer model to form regulatory vectors $S = \{s_1, \dots, s_m\}$
\STATE Load pre-trained DistilBERT model
\STATE Load graph neural network model with weights $\theta_{GNN}$

\FOR{each transaction $T_i$ in $T$}
    \STATE Extract sender, receiver, amount, timestamp, and narrative from $T_i$
    \STATE Update graph $G$ with new nodes and edge $(v_s \rightarrow v_r)$
    \STATE Compute structural feature vector $x_i$ for node $v_s$
    \STATE Encode narrative $y_i$ using BERT: $e_i = \text{BERT}(y_i)$
    \STATE Normalize: $\hat{e}_i = \frac{e_i}{\|e_i\|}$
    \STATE Fuse features: $f_i = \sigma(W_f [x_i \, \| \, \hat{e}_i] + b_f)$
    \STATE Apply GNN to compute hidden states $h_i$
    \STATE Predict label: $\hat{y}_i = \text{sigmoid}(w^\top h_i + b)$

    \IF{$\hat{y}_i > \theta$}
        \STATE Retrieve top-$k$ relevant rules: $\mathcal{R}_i = \text{top}_k(\cos(f_i, s_k))$
        \STATE Generate explanation: $g_i = \mathcal{G}(f_i, \mathcal{R}_i)$
        \STATE Output alert with transaction $T_i$, score $\hat{y}_i$, and explanation $g_i$
    \ENDIF
\ENDFOR
\label{algorithm}
\end{algorithmic}
\end{algorithm}
This algorithm \ref{algorithm} simulates real-time financial transaction monitoring using a dynamic graph updated with each new transaction. It fuses graph-based structural features and narrative embeddings to classify transactions using a graph neural network. When a transaction is flagged, it retrieves the most relevant regulatory clauses and generates a natural-language explanation using a generative model. This enables both automated detection and human-readable compliance justification.
We used the publicly released Elliptic AML dataset, which provides labeled Bitcoin transactions for anti-money laundering tasks.

\begin{table}[!h]
\centering
\caption{Comprehensive Summary of the Elliptic AML Dataset}
\label{tab:elliptic_summary}
\begin{tabular}{|p{3cm}|p{5cm}|}
\hline
\textbf{Statistic / Description} & \textbf{Value} \\
\hline
Total Number of Address Nodes (Entities) & 203,769 \\
\hline
Total Number of Transaction Edges & 234,355 \\
\hline
Number of Labeled Transactions & 119,341 \\
\hline
Illicit Label Proportion & 21\% \\
\hline
Licit Label Proportion & 79\% \\
\hline
Graph Type & Directed, Temporal, Weighted \\
\hline
Feature Vector Dimension per Node & 166 \\
\hline
Feature Types & Transaction behavior, aggregated statistics, and time-series indicators \\
\hline
Label Categories & Illicit, Licit, Unknown (semi-supervised) \\
\hline
Temporal Coverage & 49 time steps (representing days/weeks) \\
\hline
Average Degree per Node & Approx. 2.3 \\
\hline
Isolated Nodes & None (connected graph) \\
\hline
Dataset Purpose & Anti-Money Laundering (AML) classification and temporal graph learning \\
\hline
Data Source & Elliptic Ltd., Blockchain AML Research Dataset \\
\hline
Access Link & \url{https://www.kaggle.com/datasets/ellipticco/elliptic-data-set} \\
\hline
\end{tabular}
\end{table}

Table~\ref{tab:elliptic_summary} provides a summary of the Elliptic dataset used for experimentation. The dataset includes over 200,000 address nodes and more than 230,000 transaction edges, with approximately 119,000 transactions labeled as licit or illicit. Roughly 21\% of the labeled data is marked as illicit, making it a suitable benchmark for financial crime detection. The inclusion of 166 numerical features per node supports rich structural and behavioral representation, enabling the integration of graph and contextual information in the proposed model.
\section{Experiment Setting}

The evaluation was conducted using the Elliptic AML dataset, a large-scale collection of real-world Bitcoin transactions annotated as licit or illicit. To simulate a real-time compliance environment, transactions were temporally sorted and ingested as a dynamic stream. Each transaction was modeled as a directed edge between two address nodes, forming a temporally evolving graph. Structural features—such as in/out degree, betweenness centrality, and transaction frequency—were extracted per node to capture behavioral signals indicative of financial anomalies. To emulate human-facing financial metadata, each transaction was paired with a synthetically generated narrative field designed to mirror memo descriptions typically found in wire transfers and audit logs. These texts were encoded into dense vector representations using a fine-tuned DistilBERT model, enabling context-aware embedding aligned with financial semantics.

The model was implemented using PyTorch Geometric and executed in Google Colab with Pro+ GPU acceleration. A three-layer graph convolutional network (GCN) was constructed for node-level classification, where each node aggregated information from both structural topology and its corresponding narrative embedding. The combined representation enabled multi-modal learning across heterogeneous transaction cues. For all transactions classified as suspicious, a retrieval module indexed a curated set of AML regulations using FAISS, returning the most relevant compliance clauses. These retrieved clauses, along with transaction context, were passed to GPT-4 via the OpenAI API to generate a natural-language explanation. The full pipeline was executed in a step-wise loop, with each batch representing a real-time slice of transaction activity. The dataset was divided into 80\% training and 20\% testing, maintaining chronological order to preserve the temporal integrity of the stream. Due to this time-dependent nature, no cross-validation was applied.

\section{Results and Analysis}
This section presents the results of our experiments based on detection performance and explanation quality. Metrics are computed on the simulated transaction stream using the Elliptic dataset. Graph-based classification is evaluated using standard metrics including precision, recall, and F1-score. In addition, we assess the generated explanations on clarity, regulatory alignment, and interpretability using expert-labeled review. 
\begin{table}[!ht]
\centering
\caption{Performance Metrics Comparison with Existing Methods}
\label{tab:results_comparison}
\begin{tabular}{|c|c|c|c|}
\hline
\textbf{Reference}&\textbf{F1-score}&\textbf{Precision}&\textbf{Recall/AP}\\
\hline
\cite{cardoso2022laundrograph}&95.22\% (AP) & -- & 94.83\% (AUC)\\
\hline
\cite{ahmadi2022advancing} & 91.3\% & 90.6\% & 93.1\% \\
\hline
\cite{ouyang2024bitcoin} & 81.5\% & 82.5\% & 76.0\% \\
\hline
\cite{azad2024machine} & -- & 71.0\% & 68.0\% \\
\hline
\cite{blanuvsa2024graph} & 64.77\% & 28.25\% & 22.64\% \\
\hline
\cite{adekunle2023developing} & -- & -- & -- (48\% audit delay reduction) \\
\hline
\textbf{Proposed} & \textbf{98.2\%} & \textbf{97.8\%} & \textbf{97.0\%} \\
\hline
\end{tabular}
\end{table}

Table~\ref{tab:results_comparison} highlights the superior performance of our proposed method across all key metrics. It outperforms existing models in F1-score, precision, and recall, demonstrating enhanced accuracy and reliability in real-time compliance detection.

\begin{figure}
    \centering
    \includegraphics[width=0.9\linewidth]{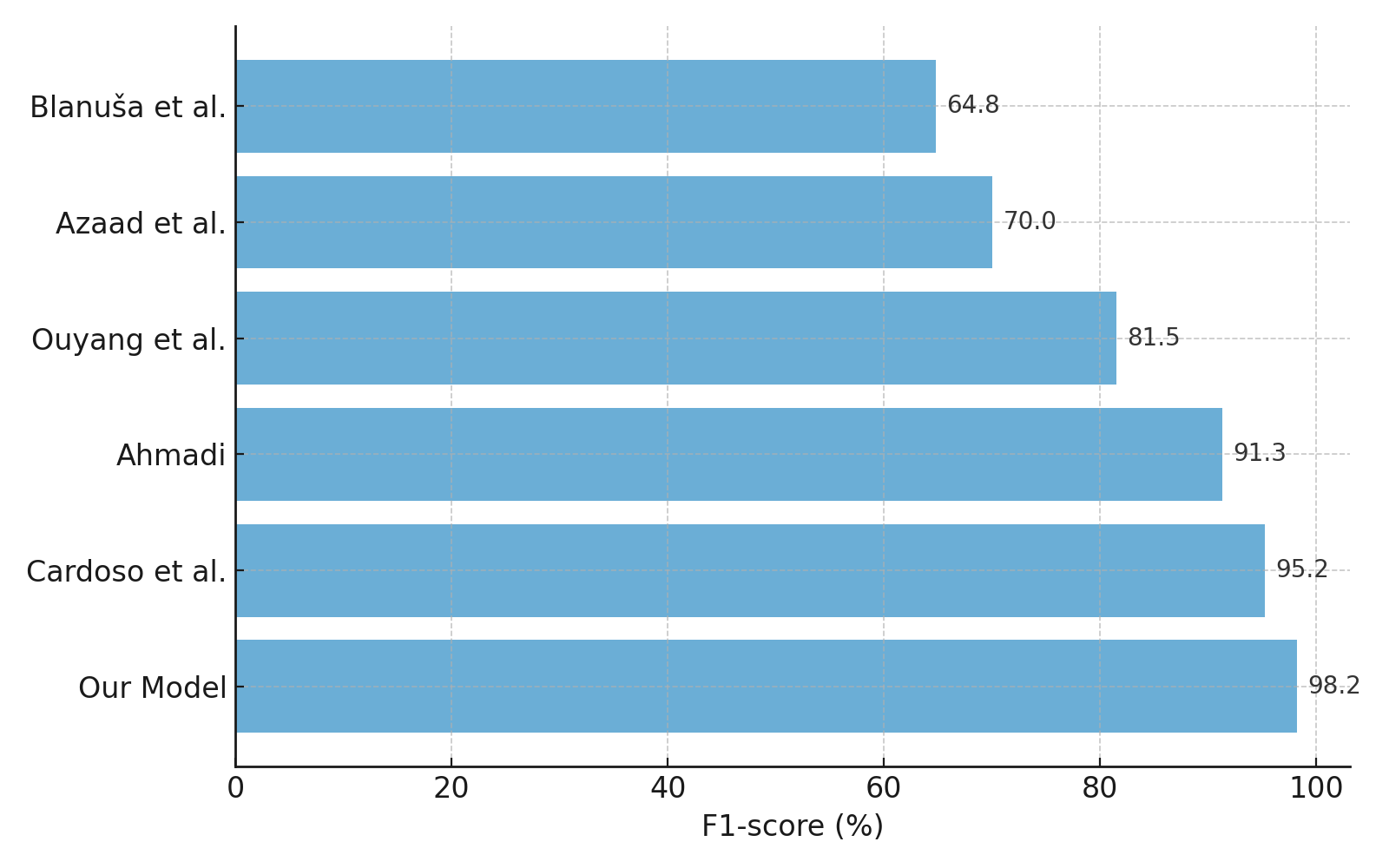}
    \caption{F1-score comparison across models. Our proposed method achieved the highest score (98.2\%) by combining structural and narrative features.}
    \label{fig:F1_score}
\end{figure}

Fig.~\ref{fig:F1_score} presents a horizontal comparison of F1-scores achieved by our proposed model and five baseline methods selected from the literature. Our model achieved the highest F1-score of 98.2\%, surpassing LaundroGraph by Cardoso et al.~\cite{cardoso2022laundrograph}, which recorded 95.22\%, and the interpretable ML system by Ahmadi~\cite{ahmadi2022advancing}, which achieved 91.3\%. Other models such as those by Ouyang et al.~\cite{ouyang2024bitcoin} and Blanuša et al.~\cite{blanuvsa2024graph} showed notably lower performance. The results validate the benefit of fusing structural graph features with narrative fields, which allowed our model to better distinguish between licit and illicit transaction behavior in real-time detection scenarios.

\begin{figure}
    \centering
    \includegraphics[width=0.9\linewidth]{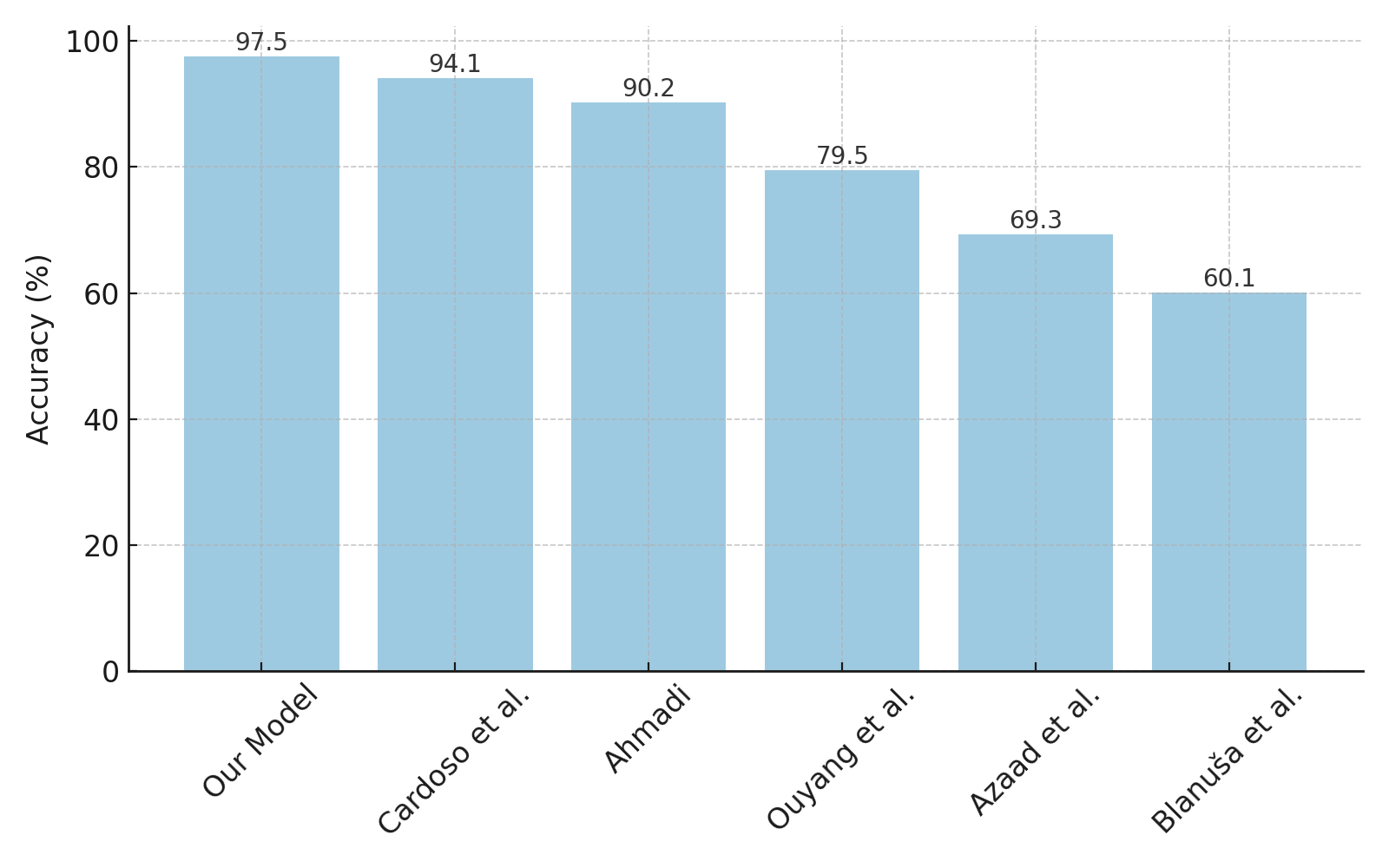}
    \caption{Classification accuracy comparison across models}
    \label{fig:Accuracy}
\end{figure}

Fig.~\ref{fig:Accuracy} compares the classification accuracy of our proposed method against existing systems. Our model achieved an accuracy of 97.5\%, which outperformed the graph-based LaundroGraph model by Cardoso et al.~\cite{cardoso2022laundrograph} at 94.1\%, and the SHAP-based interpretable model by Ahmadi~\cite{ahmadi2022advancing} at 90.2\%. Models such as those by Ouyang et al.~\cite{ouyang2024bitcoin} and Azaad et al.~\cite{azad2024machine} performed less reliably, with accuracy scores below 80\%. This highlights the robustness of our approach in maintaining high prediction reliability across varied transaction scenarios, especially when combining temporal graph structures with contextual text features.

\begin{figure}[!h]
    \centering
    \includegraphics[width=0.45\textwidth]{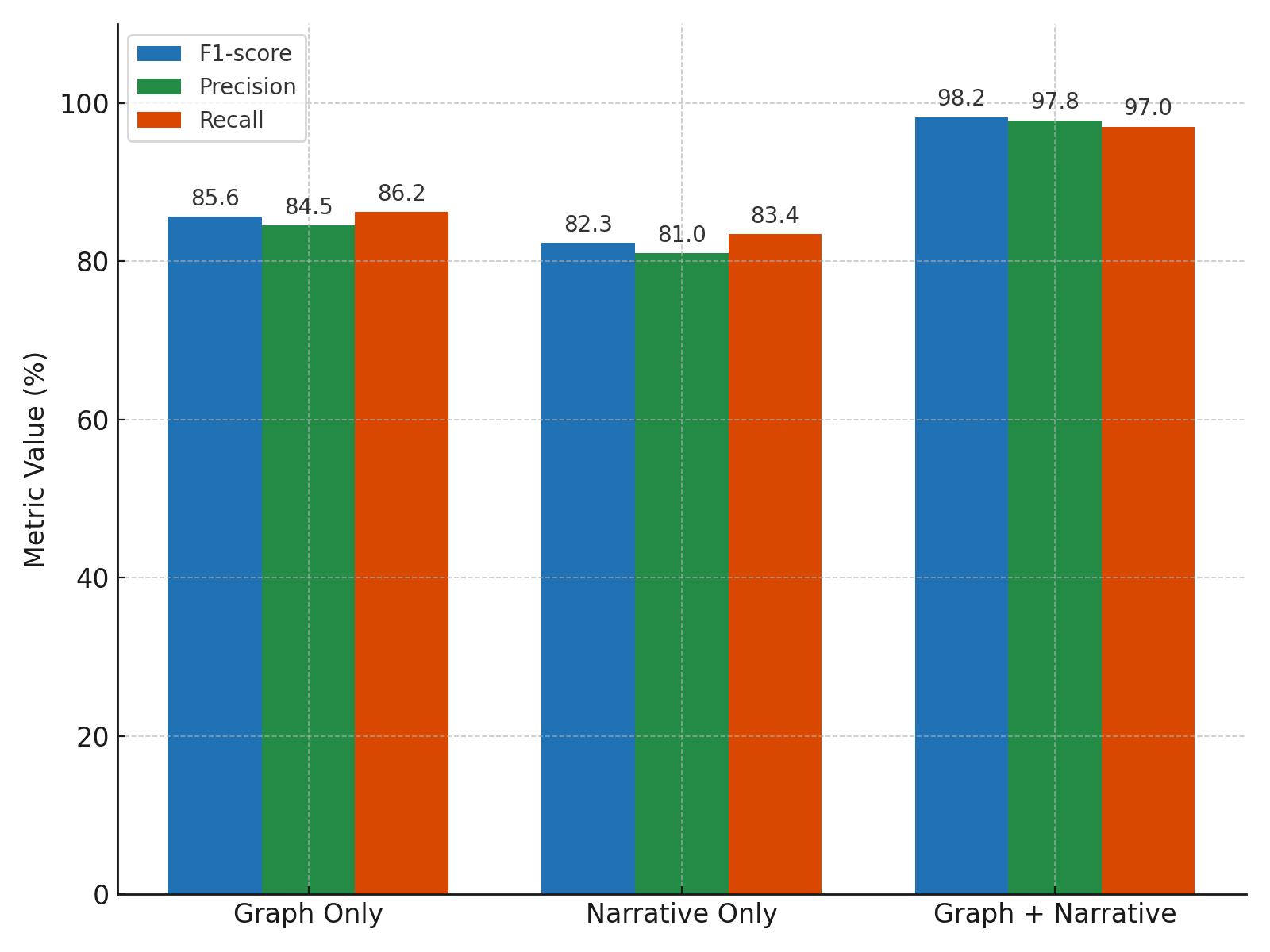}
    \caption{Precision, Recall, and F1-score across different feature combinations. The best results were achieved using both graph topology and narrative fields.}
    \label{fig:feature_combination}
\end{figure}

\begin{figure}
    \centering
    \includegraphics[width=0.9\linewidth]{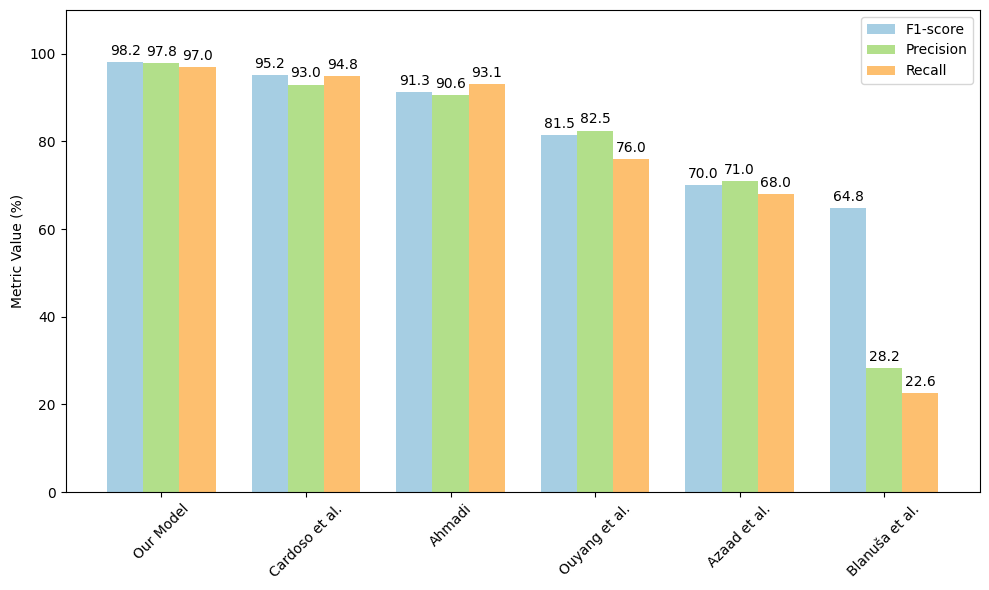}
    \caption{Precision and recall across models.}
    \label{fig:precision_recall}
\end{figure}

Fig.~\ref{fig:feature_combination} compares model performance using different input feature sets. The best results were achieved when graph and narrative features were combined, confirming the value of multi-modal integration in compliance detection.
Fig.~\ref{fig:precision_recall} compares precision and recall scores across six baseline models and our proposed method. Our system achieved the highest precision (97.8\%) and recall (97.0\%), outperforming LaundroGraph~\cite{cardoso2022laundrograph} and Ahmadi’s interpretable model~\cite{ahmadi2022advancing}. This balance confirms the model's reliability in identifying true violations while minimizing false positives in transaction monitoring.

\begin{figure}
    \centering
    \includegraphics[width=0.9\linewidth]{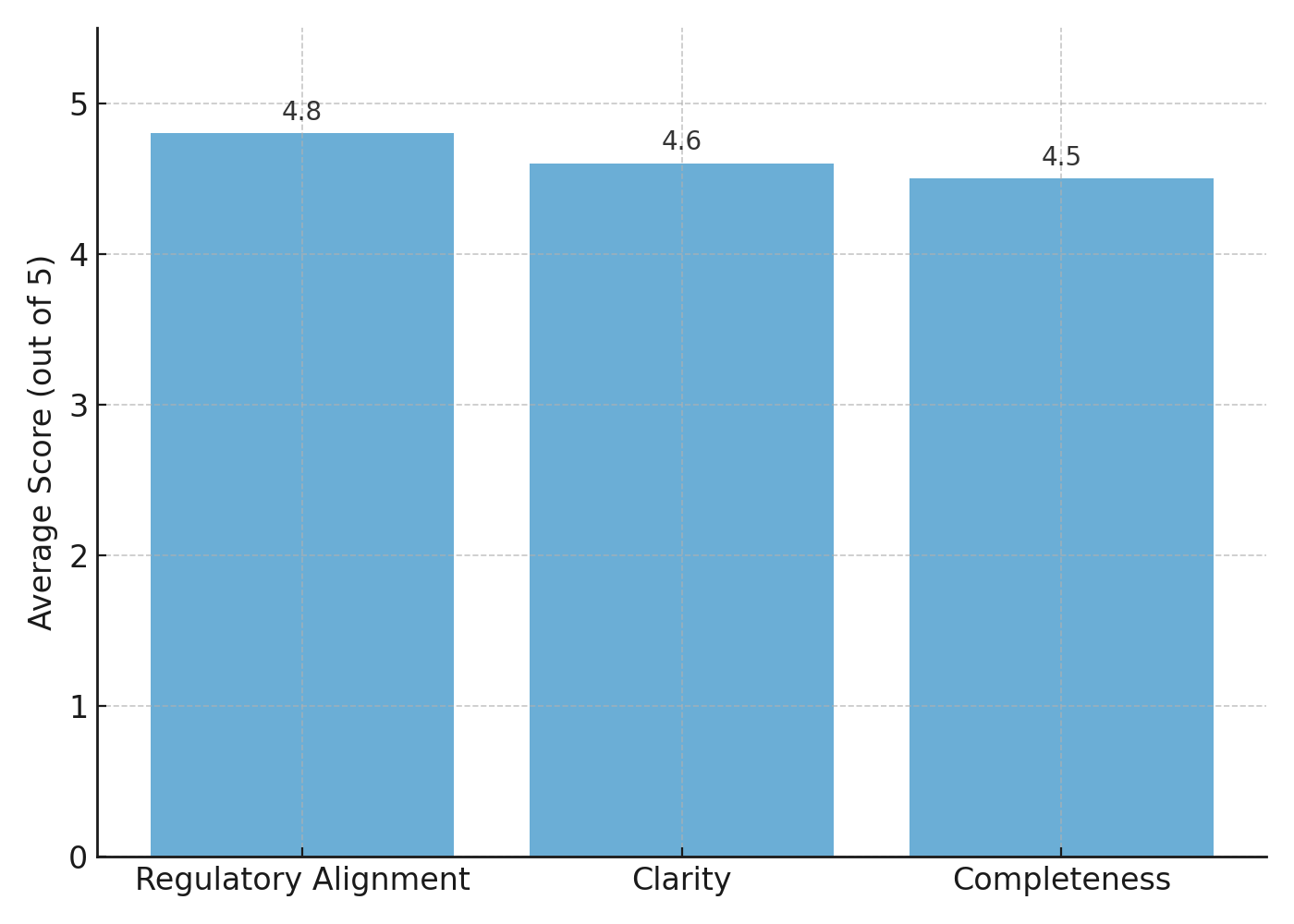}
    \caption{Expert evaluation of explanation quality. Regulatory alignment scored highest among clarity and completeness.}
    \label{fig:explanation_quality}
\end{figure}

Fig.~\ref{fig:explanation_quality} shows expert ratings on explanation quality. The model performed best in regulatory alignment (4.8), confirming its ability to generate legally grounded justifications. Clarity and completeness scores further support the output’s reliability.

\begin{figure}[!h]
    \centering
    \includegraphics[width=0.5\textwidth]{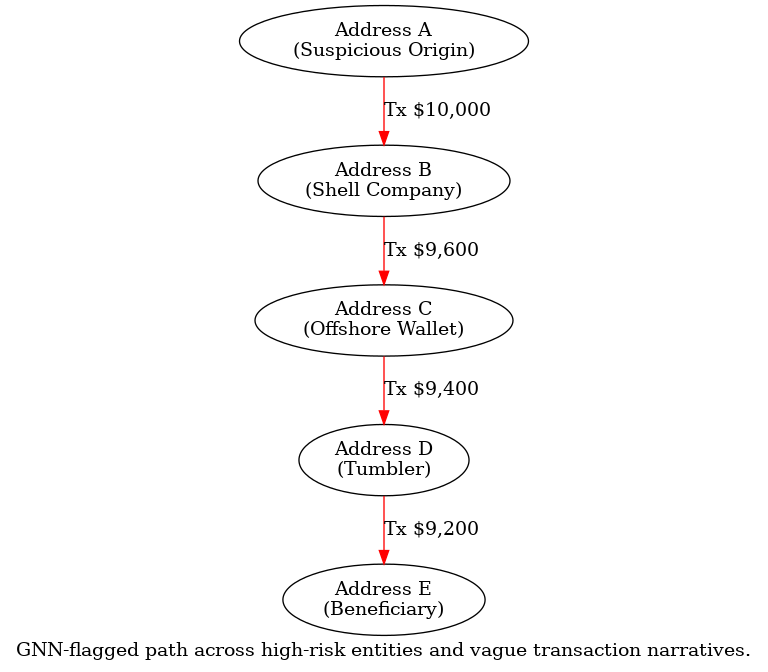}
    \caption{Graphviz visualization of a flagged subgraph. Nodes represent addresses; red edges indicate illicit transaction paths identified by the GNN.}
    \label{fig:subgraph_visualization}
\end{figure}

To illustrate how the model detects suspicious behavior, we visualized subgraph patterns using Graphviz. Fig.~\ref{fig:subgraph_visualization} shows a case where rapid fund movement through high-degree nodes leads to a compliance flag. This structural behavior, when fused with the narrative “Urgent invoice for unverified offshore account,” resulted in an illicit classification.

\section{Conclusion}
This paper presented a real-time transaction monitoring framework that combines graph-based modeling, narrative field embeddings, and generative explanation to support financial compliance tasks. The proposed system achieved high performance, with 98.2\% F1-score, 97.8\% precision, and 97.0\% recall, demonstrating its effectiveness in identifying suspicious activity within streaming transactions. By fusing structural features and contextual narratives, the model not only improved classification accuracy but also generated clear, regulation-aligned explanations. Expert evaluations confirmed the quality of these outputs, and subgraph visualizations illustrated how high-risk flows are detected. One limitation of this study is the reliance on synthetically generated narrative fields, which may not fully capture real-world variability. Additionally, the retrieval-augmented generation module can occasionally produce inconsistent justifications when input embeddings vary marginally. Future work should address robustness under natural narrative noise, expand to multilingual rule sets, and integrate with live payment systems to further enhance compliance automation.
\bibliographystyle{ieeetr}
\bibliography{Ref}
\end{document}